\documentclass[letterpaper, 10 pt, journal]{ieeetran}  

\usepackage{bm}
\usepackage{amsmath} 
\usepackage{amssymb}  
\usepackage{lipsum}
\usepackage{xcolor}
\usepackage{algorithm}
\usepackage[]{algpseudocode}
\usepackage{graphicx}
\usepackage[hyphens]{url}
\usepackage[hidelinks]{hyperref}
\usepackage{hyperref}
\usepackage{cite}
\usepackage{array}
\usepackage{caption}
\usepackage{subcaption}
\usepackage{multirow}
\usepackage{siunitx}
\usepackage{outlines}

\usepackage{url}
\usepackage{tikz}
\usepackage{tikz-3dplot}
\usetikzlibrary{shapes,arrows}
\usetikzlibrary{positioning}

\usepackage[font=footnotesize,labelfont=bf]{caption}

\title{\LARGE \bf
Range-Visual-Inertial Sensor Fusion for Micro Aerial Vehicle Localization and Navigation 
}

\author{Abhishek Goudar, Wenda Zhao, and Angela P. Schoellig
\thanks{The authors are with the Learning Systems and Robotics Lab
	(www.learnsyslab.org) at the Technical University of Munich, Germany, and
	the University of Toronto Institute for Aerospace Studies, Canada. They are
	also affiliated with the University of Toronto Robotics Institute, the Munich
	Institute of Robotics and Machine Intelligence (MIRMI), and the Vector Institute for Artificial Intelligence. E-mail: \{abhishek.goudar, wenda.zhao\}@robotics.utias.utoronto.ca,	
	angela.schoellig@tum.de}%
}

\begin{document}

\maketitle
\thispagestyle{empty}
\pagestyle{empty}

\begin{abstract}


We propose a fixed-lag smoother-based sensor fusion architecture to leverage the complementary benefits of range-based sensors and visual-inertial odometry (VIO) for localization. We use two fixed-lag smoothers (FLS) to decouple accurate state estimation and high-rate pose generation for closed-loop control. The first FLS combines ultrawideband (UWB)-based range measurements and VIO to estimate the robot trajectory and any systematic biases that affect the range measurements in cluttered environments. The second FLS estimates smooth corrections to VIO to generate pose estimates at a high rate for online control. The proposed method is lightweight and can run on a computationally constrained micro-aerial vehicle (MAV). We validate our approach through closed-loop flight tests involving dynamic trajectories in multiple real-world cluttered indoor environments. Our method achieves decimeter-to-sub-decimeter-level positioning accuracy using off-the-shelf sensors and decimeter-level tracking accuracy with minimally-tuned open-source controllers.

\end{abstract}

\newcommand{\state}{\mathbf{T}^W_{it}}

\newcommand{\pose}{\mathbf{T}}
\newcommand{\nompose}{\bar{\mathbf{T}}}
\newcommand{\perturb}{\bm{\xi}}

\section{INTRODUCTION}

Range-based positioning involves determining the position of a robot by measuring the distance to specific landmarks known as anchors. Range-based positioning is appealing as it is lightweight, drift-free, and computationally inexpensive. Additionally, unlike camera or lidar-based positioning, it does not require a detailed map of the environment or persistent features. Common approaches to range-based positioning include radio-frequency (RF)-based positioning technologies such as the Global Positioning System (GPS) for outdoor environments \cite{kaplan}

However, urban canyons and most indoor environments are not amenable to GPS-based positioning. Ultrawideband (UWB) \cite{yavari2014} is an alternative RF-based technology suitable for such environments. A challenge of RF technologies such as UWB is that the measurements are affected by objects in the environment that obstruct the line of sight between the anchor and the robot, resulting in biased \textit{non-line of sight} (NLOS) measurements \cite{Prorok2012}. The effect of NLOS measurements is more pronounced in cluttered environments. The combination of biases and the low-dimensional nature of range measurements can result in non-smooth state estimates that is undesirable for closed-loop control \cite{Hoeller2017}.

Since range measurements are low dimensional, a single range sensor cannot determine the full pose of a robot. As such, range sensors are typically combined with other sensing modalities such as wheel odometry, inertial odometry, and visual odometry \cite{hol2009, blanco2008, Nguyen2021}. Decreasing cost of high-quality cameras and inertial measurement units (IMUs) has led to the development of mature visual-inertial odometry (VIO) \cite{Scaramuzza2020} algorithms. VIO provides smooth relative 3D pose measurements at a high rate. We refer to output obtained from a VIO algorithm as simply odometry. The smooth nature of VIO makes it suitable for closed-loop control. However, VIO is susceptible to drift, especially in dynamic environments with changing scenery.

\begin{figure}
	\centering
	\raisebox{-0.75ex}{\includegraphics[clip,trim=3.cm  2cm 2.6cm 3cm, width=0.24\textwidth]{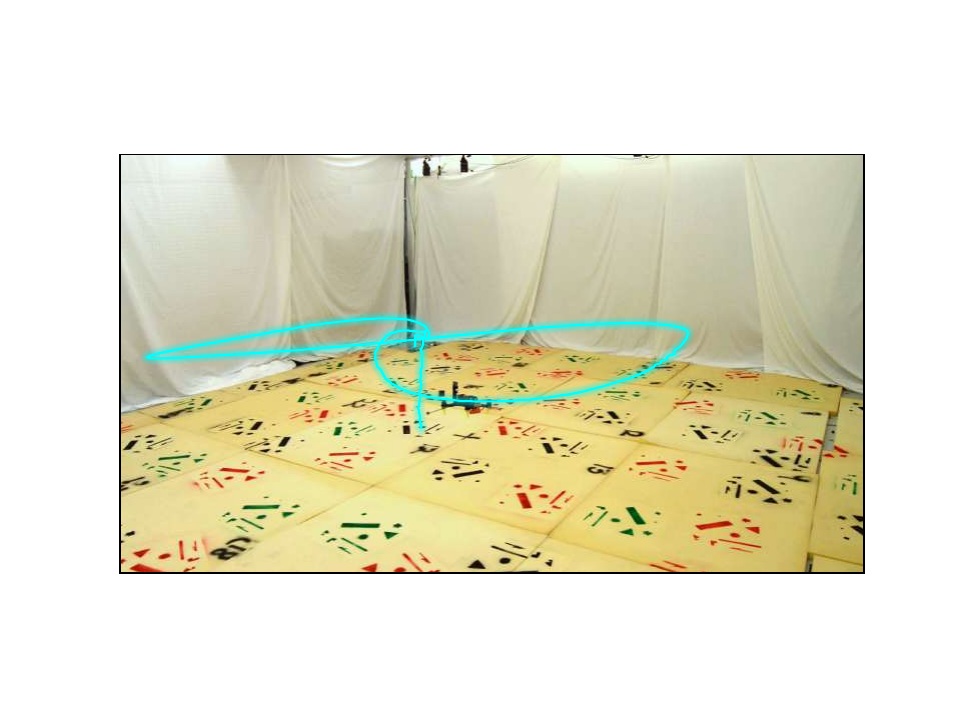}}
	\vspace*{-1.em}
	\hspace*{-1.em}
	\includegraphics[clip,trim=2.35cm 2cm 2.55cm 2.85cm, width=0.24\textwidth]{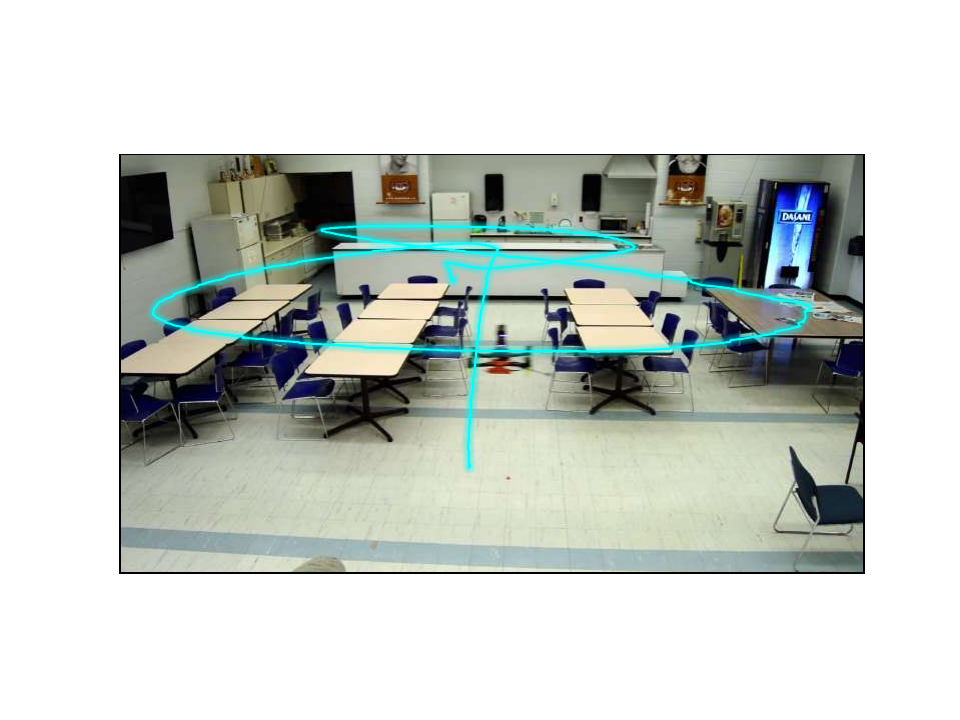}
	\\
	\includegraphics[clip,trim=2.5cm 3.0cm 3cm 2.7cm, width=0.24\textwidth]{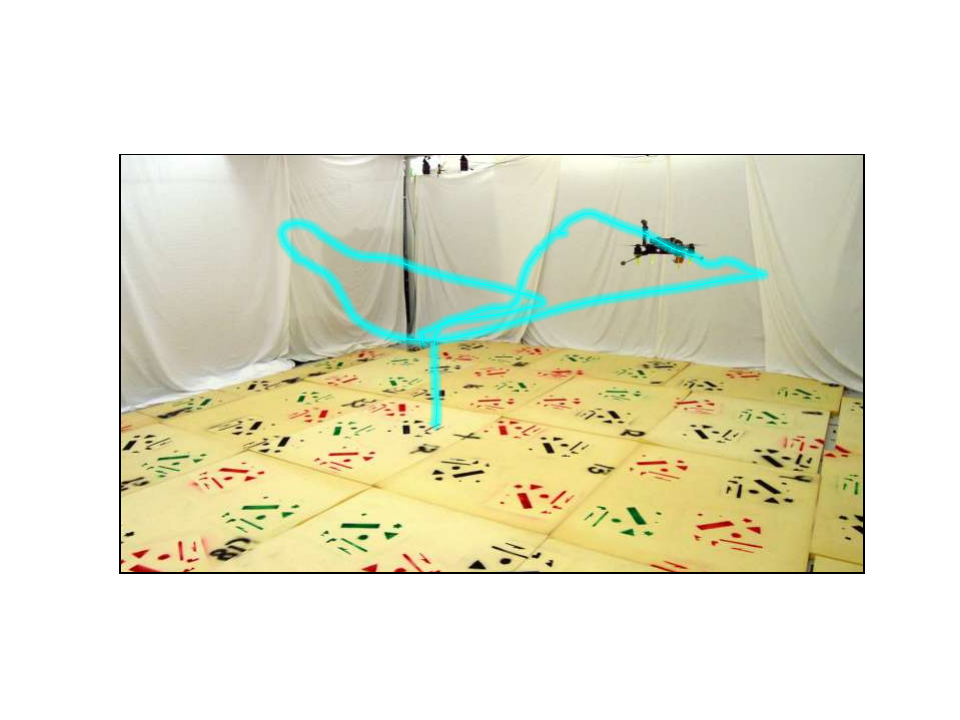}
	\hspace*{-1.em}
	\includegraphics[clip,trim=2.5cm 3.0cm 3cm 2.7cm, width=0.24\textwidth]{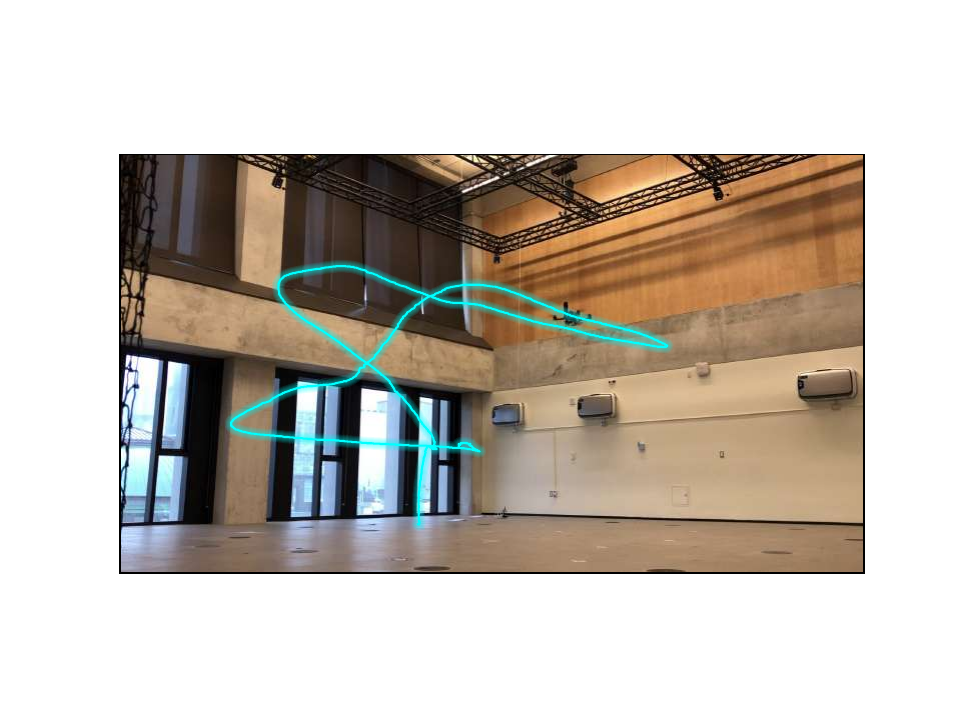}
	\caption{Overlay of trajectories from closed-loop flight experiments using our proposed method for localization. The proposed method uses a dual-rate fixed-lag smoother architecture to combine range measurements from ultrawideband radios and visual inertial odometry for localization. The UTIAS testbed (left top and left bottom) and the Myhal testbed (right bottom) present challenging scenarios in terms of poor geometry of anchors for range-based positioning, and sparse features for visual inertial odometry, respectively. The UTIAS cafeteria (right top) is challenging as ultrwideband signals are affected by reflections from the obstacles.  A video of our experiments is available at: {\label{video}\url{http://tiny.cc/uwb_vio}}.}
	\label{fig:myhal_motion_shot}
	\vspace*{-1em}
\end{figure}

The above discussion highlights the complementary nature of range-based positioning and odometry-based positioning. We refer to the combination of range and odometry-based localization as \textit{range-aided} localization. To leverage their complementary benefits for accurate state estimation \textit{and} closed-loop control, any sensor fusion scheme must
\textit{(i)} be tractable for online estimation on computationally constrained platforms,
\textit{(ii)} estimate any biases associated with the range measurements, and
\textit{(iii)} provide high-rate, drift-free, smooth state estimates for reliable trajectory tracking.

A common approach to state estimation is to use a \textit{batch} trajectory estimator \cite[Chapter 4]{Barfoot2023}. While such an approach can provide smooth and accurate state estimates, it is not well-suited for online control. Filtering-based methods \cite{hol2009, Mueller2015, Hausman}, while computationally lightweight, can result in poorer accuracy and non-smooth estimates as they discard older measurements. These non-smooth estimates can result in oscillatory behavior when used for online control \cite{Heijden}. In this paper, we propose a sensor fusion architecture that meets all of the above requirements. The following are the main contributions of our work.
\begin{outline}
	\1 We propose a dual fixed-lag smoother (FLS) architecture to fuse range measurements from UWB radios and VIO for localization and navigation. The decoupled nature of our architecture enables high-accuracy trajectory estimation and high-rate smooth pose generation for online control simultaneously.
	\1 We present a computationally lightweight method to estimate the systematic biases in range measurements. Unlike previous works, our approach does not require ground truth pose information, additional sensors, or training data. The proposed bias estimation method reduces the trajectory estimation error by $50\,\%$ compared to not estimating the biases. 
	\1 We demonstrate the effectiveness of our approach in closed-loop flights where the proposed method achieves decimeter-to-sub-decimeter-level localization accuracy and decimeter-level tracking accuracy in real-world cluttered indoor environments.
	\1 We release our source code\footnote[1]{\url{https://github.com/utiasDSL/ra_lan.git}} and the dataset\footnote[2]{\url{https://utiasdsl.github.io/utias_ra_loc}} collected during the experiments.  
\end{outline}


%

\section{RELATED WORK} \label{sec:related_work}
In this section, we review previous works that combine range sensors with other sensors to achieve accurate localization. We also review works that estimate biases in UWB measurements. Finally, we review works that focus on the fusion of range and odometry measurements for navigation.

\textbf{UWB-aided localization and navigation}: As alluded to earlier, a single UWB radio cannot estimate the full 3D pose, and as such it is used with other sensors such as an IMU or camera. The fusion of UWB and IMU measurements can be done either by using range measurements, referred to as \textit{range-based approach}, or by first estimating position from range measurements and then using the position as input, referred to as \textit{position-based approach}. Systems adopting range-based approaches \cite{hol2009,goudar2021}, position-based approaches \cite{guo2016}, and a combination thereof \cite{multilateration} have been shown previously. The common sensor fusion strategies include parametric filtering-based methods such as the extended Kalman filter (EKF) \cite{hol2009,Mueller2015,Hausman,Li2018,goudar2021} and non-parametric methods such as particle filters \cite{Prorok2012}. More recently, optimization-based approaches have gained traction \cite{Fang2021,Nguyen2021}. Computer vision algorithms have also been used in conjunction with UWB sensors to achieve accurate localization using filtering-based \cite{Nyqvist2015, Hausman, Hoeller2017} and optimization-based approaches \cite{Nguyen2021}. More recently, tightly-coupled approaches that combine range measurements with data from cameras, IMUs, and lidars \cite{nguyen2021viral,Cao2020VIRSLAMVI} for multi-modal simultaneous localization and mapping (SLAM) have been demonstrated. In addition to localization, our proposed system is also designed for reliable closed-loop trajectory tracking on computationally constrained platforms. 

\textbf{Bias estimation in UWB-based localization}: As mentioned previously, in cluttered environments, UWB measurements are susceptible to noise and bias. The identification of biases in UWB measurements requires additional information from other sources. Measurements from a motion capture system are used to estimate biases in \cite{guo2016,Ledergerber2017,wenda2022}. Fiducial markers-based pose estimation is used to facilitate bias estimation in \cite{Hoeller2017}. In \cite{nguyen2021viral} additional sensors such as lidars are used. The methods proposed in \cite{Ledergerber2018a,Heijden} use training data collected from multiple closed-loop flights to learn the bias models. A variety of models including parametric models \cite{Prorok2012,guo2016,Heijden} and non-parametric models such as Gaussian processes \cite{Ledergerber2017}, neural networks \cite{wenda2022}, and kernel density-based models \cite{Haggenmiller2019} have been used to estimate the biases in range measurements. Some of the limitations of previous works are that they require accurate ground truth information or training data and can be computationally intractable for online estimation. In contrast, our method does not require ground truth or training data and is tractable to run online on computationally constrained platforms.

\textbf{Multisensor fusion}: We review works that focus on the fusion range measurements obtained from UWB or GPS-based systems and odometry obtained from VIO or IMU for closed-loop control. A common approach is to combine absolute and relative measurements \textit{directly} using a filtering-based method \cite{Hausman,Lynen2013} or an optimization-based method \cite{Nguyen2021}. An alternative approach is to model the sensor fusion problem as a frame alignment problem \cite{Mascaro2018}. The benefit of such an approach is that the core estimation pipeline runs at a lower rate to correct the drift in the high-rate relative measurements, which is then used to perform control. We extend the approach of \cite{Mascaro2018} to use range measurements (instead of position measurements) with online bias estimation. Additionally, we augment the frame alignment method of \cite{Mascaro2018} to generate \textit{smooth} frame alignment.

To the best of the authors' knowledge, sensor fusion of UWB-based range measurements and odometry from VIO for \textit{(i)} estimation of the robot state, \textit{(ii)} generation of smooth state estimates at high-rate for closed-loop control, and \textit{(iii)} online estimation of systematic biases in range measurements without additional sensors on computationally constrained platforms has not been demonstrated previously.

%

\section{Problem Statement} \label{sec:prob_state}

The objective of our work is to not only estimate the robot trajectory and the biases associated with range measurements, but also to generate smooth state estimates at high-rate for closed-loop control. We assume that \textit{(i)} the position of the anchors is known, and \textit{(ii)} the robot is equipped with a UWB radio and a high-rate VIO sensor.

We introduce the notation that will be used throughout the paper. Elements of the special Euclidean \textit{Lie} group $\mathbf{T} \in {SE}(3)$ are used to represent 3D poses. A generic pose $\mathbf{T}$ is parameterized as $\mathbf{T} = \{ \mathbf{p}, \mathbf{R} \}$, where $\mathbf{p} \in \mathbb{R}^3$ represents the 3D position and $\mathbf{R} \in SO(3)$, a member of the special orthogonal group, represents the orientation. The pose of a frame $A$ in frame $B$ is represented by $\mathbf{T}^A_B = \{ \mathbf{p}^A_B, \mathbf{R}^A_B \}$. 

We denote the world frame by $\{W\}$, the robot frame by $\{i\}$, and the local frame associated with VIO by $\{o\}$. Unlike the world frame, the local frame is not fixed and is set to the point in space where the VIO algorithm is initialized. The UWB range measurements are reported in the world frame whereas odometry from VIO is in the local frame. The bias associated with the measurements from the $l^\text{th}$ anchor, $a_l$, is denoted by $b_{a_l} \in \mathbb{R}$. 

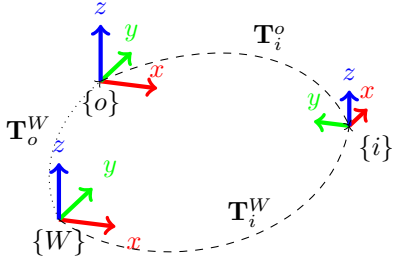
\begin{figure}[t]
	\centering
	\tdplotsetmaincoords{70}{110}
	\begin{tikzpicture}[scale=1.,tdplot_main_coords] 
		\coordinate (WORLD_ORIGIN) at (0,0,0);
		\coordinate (VIO_ORIGIN) at (2.5,1.5,3.);
		\coordinate (IMU_ORIGIN) at (3,5.2,3.);

		\node [anchor=north] at (WORLD_ORIGIN){$\{W\}$};
		\node [anchor=north west] at (IMU_ORIGIN){$\{i\}$};
		\node [anchor=north] at (VIO_ORIGIN){$\{o\}$};

		\draw[ultra thick,->, red] (WORLD_ORIGIN) -- (0,0.8,0) node[anchor=north west]{$x$};
		\draw[ultra thick,->, green] (WORLD_ORIGIN) -- (-1.3,0,0) node[anchor=south west]{$y$};
		\draw[ultra thick,->, blue] (WORLD_ORIGIN) -- (0,0,0.8) node[anchor=south]{$z$};

		\tdplotsetrotatedcoords{0}{0}{0}
		\tdplotsetrotatedcoordsorigin{(VIO_ORIGIN)}
		\draw[ultra thick,color=red,tdplot_rotated_coords,->] (0,0,0) --(0,0.8,0) node[anchor=south]{$x$};
		\draw[ultra thick,color=green,tdplot_rotated_coords,->] (0,0,0) --(-1.2,0,0) node[anchor=south]{$y$};
		\draw[ultra thick,color=blue,tdplot_rotated_coords,->] (0,0,0) --(0,0,0.8) node[anchor=south]{$z$};

		\tdplotsetrotatedcoords{0}{0}{0}
		\tdplotsetrotatedcoordsorigin{(IMU_ORIGIN)}
		\draw[ultra thick,color=red,tdplot_rotated_coords,->] (0,0,0) --(-.7,0,0) node[anchor=south]{$x$};
		\draw[ultra thick,color=green,tdplot_rotated_coords,->] (0,0,0) --(0,-0.5,0) node[anchor=south]{$y$};
		\draw[ultra thick,color=blue,tdplot_rotated_coords,->] (0,0,0) --(0,0,0.5) node[anchor=south]{$z$};

		\draw[dashed, black, ->] (WORLD_ORIGIN) to [out=325, in=260] node[above, pos=0.55]{${\mathbf{T}}^W_i$} (IMU_ORIGIN);
		\draw[dashed, black, ->] (VIO_ORIGIN) to [out=25, in=115] node[above, pos=0.6]{$\mathbf{T}^o_i$} (IMU_ORIGIN);
		\draw[dotted, black, ->] (WORLD_ORIGIN) to [out=115, in=220] node[left, pos=0.6]{$\mathbf{T}^W_o$} (VIO_ORIGIN);

	\end{tikzpicture}
	\caption{The problem of sensor fusion of UWB range measurements and VIO odometry is modeled as a frame alignment problem between the world frame $\{W\}$ and the odometry frame $\{o\}$. The \textit{absolute} but non-smooth estimate $\mathbf{T}^W_i$ obtained from fusing UWB and VIO measurements is combined with the \textit{relative} but smooth estimate ${\mathbf{T}}^o_i$ from VIO to estimate the offset $\mathbf{T}^W_o$ between the two frames. }
	\label{fig:frame_setup}
	\vspace*{-1em}
\end{figure}

\section{MODELLING} \label{sec:modelling}

In the next subsection, we provided an overview of our system architecture. A detailed description of the individual components is provided in the subsequent subsection.

\subsection{Motivation and System description} \label{sec:motivation}

Similar to \cite{Mascaro2018}, we formulate the problem of sensor fusion as a frame alignment problem. Unlike \cite{Mascaro2018}, which considers a position-based approach, we use a range-based approach. A limitation of the position-based approach is that the effects of the biases in range measurements are not discernible from the position measurements. In a range-based approach, biases result in longer range measurements, and hence, the biases can be estimated.

\begin{figure}[t]
	\centering
	\includegraphics[scale=0.48]{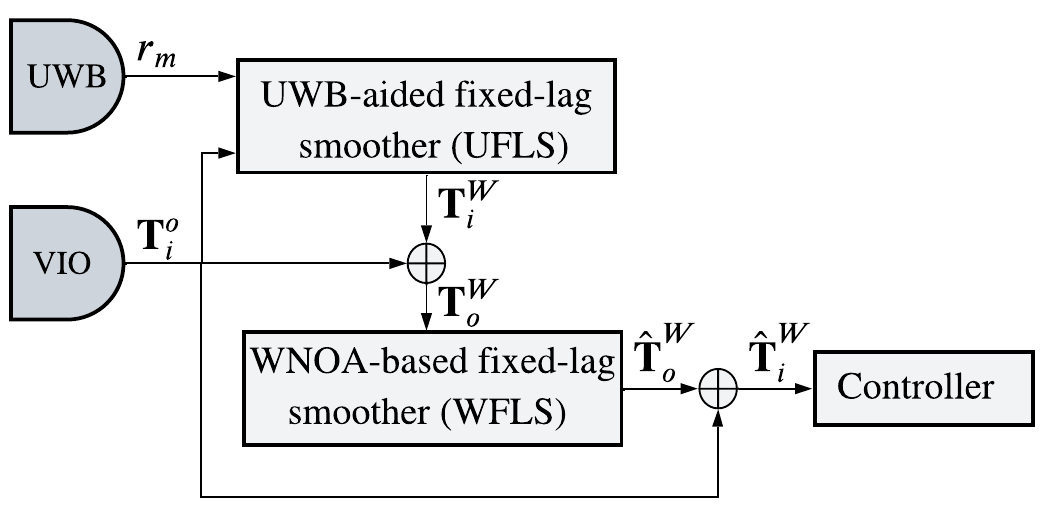}
	\caption{Overview of the proposed architecture. Range measurements from UWB radios and odometry from VIO are combined in the first fixed-lag smoother (UFLS) to estimate the robot pose ${\mathbf{T}}^W_i$ and the systematic biases in range measurements. The estimated robot pose is then combined with the latest estimate from VIO $\mathbf{T}^o_i$ to estimate the frame offset  ${\mathbf{T}}^W_o$ between the world frame and the local frame. A white-noise-on-acceleration (WNOA) motion model is used to generate a smooth frame offset, $\hat{\mathbf{T}}^W_o$, using the second fixed-lag smoother (WFLS). The smoothed frame offset is then combined with VIO-baed odometry to obtain smooth, high-rate, and drift-free robot pose $\hat{\mathbf{T}}^W_i$ which is sent to the controller.}
	\label{fig:sys_arch}
	\vspace*{-1em}
\end{figure}

The block diagram of our system architecture is shown in Figure \ref{fig:sys_arch}. First, we combine range measurements and odometry to estimate the robot trajectory and any systematic biases in range measurements using a fixed-lag smoother (UFLS). The estimated robot pose, $\mathbf{T}^W_i$, is non-smooth due to the sparse and non-smooth nature of range measurements. To keep the computational cost low, UFLS is run at a lower rate compared to individual sensor update rates. However, the low-rate and non-smooth nature of the output of UFLS makes it unsuitable for online control.

The output of VIO, $\mathbf{T}^o_i$, is typically high-rate and smooth but drifts over time. An approach to get high-rate drift-free estimates is to combine the output of UFLS, $\mathbf{T}^W_i$, and the output of VIO, $\mathbf{T}^o_i$, by performing frame alignment. Specifically, the output of UFLS and VIO can be used to obtain the frame offset between the world frame and the local frame
\begin{equation}
\mathbf{T}^W_{o} = \mathbf{T}^W_i (\mathbf{T}^o_i)^{-1}.
\label{eqn:frame_offset}
\end{equation}
The frame diagram for such a setup is shown in Figure \ref{fig:frame_setup}. We have omitted the dependency on time for notational clarity. This frame offset can be combined with the latest odometry-based pose estimate \cite{Mascaro2018} to generate high-rate and drift-free pose estimates, $\tilde{\mathbf{T}}^{W}_{i} = \mathbf{T}^W_o \mathbf{T}^o_i$.
While this addresses the issue of generating drift-free high-rate pose estimates, in a range-based setup, the frame offset $\mathbf{T}^W_{o}$ (and hence $\tilde{\mathbf{T}}^{W}_{i}$) is non-smooth on account of $\mathbf{T}^W_i$ being non-smooth.

To generate smooth frame offset estimates, we impose a white-noise-on-acceleration (WNOA) \cite{Barfoot2014a, Anderson2015} motion model on the frame offset. The motivation for choosing such a model is that \textit{(i)} the WNOA model provides a \textit{smooth prior} which acts as a regularization term to filter any non-smooth component, and \textit{(ii)} the WNOA model leads to motion priors with sparse system matrices which be solved efficiently. The motion prior and the frame offset estimates obtained from \eqref{eqn:frame_offset} are combined in a WNOA-based fixed-lag smoother (WFLS in Figure \ref{fig:sys_arch}) to generate smooth frame offset $\hat{\mathbf{T}}^W_{o}$. This smoothed frame offset is then combined with the latest odometry-derived pose estimate to generate drift-free, high-rate, and smooth robot pose estimate, $\hat{\mathbf{T}}^W_i = \hat{\mathbf{T}}^W_o \mathbf{T}^o_i$. 

The advantage of such an architecture is that the estimation of the biases and the robot trajectory can happen at a lower rate while still being able to generate smooth state estimates using drift-corrected odometry for online control. Note that the odometry from VIO is used twice in entire pipeline. However, the odometry estimate used for calculating the frame offset is different (older) compared to the odometry estimate combined with the smoothed frame offset for control, which precludes double counting of odometry.

\subsection{UWB-aided fixed-lag smoother}

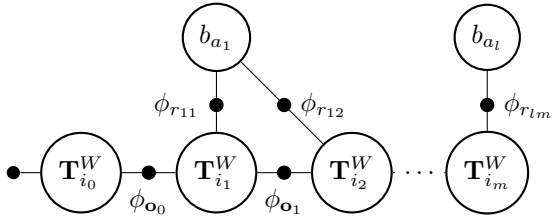
\begin{figure}[!t]
	\center
	\begin{tikzpicture}[
	node distance={9mm},
	vertex/.style={circle, draw=black!100, thick, minimum width=0.85cm},
	empty_factor/.style={minimum size=1mm},
	]
	\node[circle,fill=black,inner sep=0pt,minimum size=5pt] (prior) at (0,0) {};
	\node[vertex] (x0) [right of=prior] {$\mathbf{T}^W_{i_0}$};
	\node[circle, draw, right of=x0, fill=black, inner sep=0pt,minimum size=5pt, label=below:{$\phi_{\mathbf{o}_0}$}] (o0){};
	\node[vertex] (x1)  [right of=o0] {$\mathbf{T}^W_{i_1}$};
	\node[circle, draw, right of=x1, fill=black, inner sep=0pt,minimum size=5pt, label=below:{$\phi_{\mathbf{o}_1}$}] (o1){};
	\node[vertex] (x2)  [right of=o1] {$\mathbf{T}^W_{i_2}$};
	\node[empty_factor] (o2) [right of=x2] {$\hdots$};
	\node[vertex] (xM) [right of=o2] {$\mathbf{T}^W_{i_m}$};
	\node[circle, draw, above of=x1, fill=black, inner sep=0pt,minimum size=5pt, label=left:{$\phi_{r_{11}}$}] (y1){};
	\node[circle, draw, above of=o1, fill=black, inner sep=0pt,minimum size=5pt, label=right:{$\phi_{r_{12}}$}] (y2){};
	\node[circle, draw, above of=xM, fill=black, inner sep=0pt,minimum size=5pt, label=right:{$\phi_{r_{lm}}$}] (yM){};
	\node[vertex] (xa) [above of=y1] {$b_{a_1}$};
	\node[vertex] (xb) [above of=yM] {$b_{a_l}$};
	
	%
	\draw (prior) -- (x0);	
	\draw (x0) -- (o0);
	\draw (o0) -- (x1);
	\draw (x1) -- (o1);
	\draw (o1) -- (x2);
	\draw (x2) -- (o2);
	\draw (o2) -- (xM);
	\draw (x1) -- (y1);
	\draw (y1) -- (xa);
	\draw (x2) -- (y2);
	\draw (y2) -- (xa);
	\draw (xM) -- (yM);
	\draw (yM) -- (xb);
	%
	\end{tikzpicture}
	\caption{Factor graph for the UWB-aided fixed-lag smoother (UFLS). Preintegrated odometry is added as a binary factor $\phi_{\mathbf{o}_t}$ to constrain consecutive robot poses. Each range measurement adds a binary factor $\phi_{\mathbf{r}_{lt}}$ between the robot pose $\mathbf{T}^W_{i_t}$ and the node $b_{a_l}$ representing the bias associated with anchor $a_l$.}
	\label{fig:smoother_factor_graph}
	\vspace{-1em}
\end{figure}

In UFLS, a fixed window of range and odometry measurements are combined to estimate the robot trajectory and the biases in range measurements. The size of the window is parameterized by time duration $\delta t_\textrm{ufls}$. Measurements older than $\delta t_\textrm{ufls}$ are discarded and states older than $\delta t_\textrm{ufls}$ are marginalized. The factor graph corresponding to such a setup is shown in Figure \ref{fig:smoother_factor_graph}. The UWB range measurement at any time $t$ between the robot and an anchor $a_l$ is given by
\begin{equation}
	r_{lt} = \| \mathbf{p}^W_{a_l} - \mathbf{R}^W_{it} \mathbf{p}^i_u - \mathbf{p}^W_{it} \|_2 + \mathbf{b}_{a_l} + \eta_{rt},
	\label{eqn:uwb_model}
\end{equation}
where $\|\cdot\|_2$ is the $\ell^2$ norm, $\mathbf{T}^W_{it} = \{\mathbf{p}^W_{it}, \mathbf{R}^W_{it} \}$ is the pose of the robot at time $t$, $\mathbf{p}^i_u$ is the position of the UWB radio w.r.t the body frame a.k.a \textit{lever arm}, $\eta_{rt} \sim \mathcal{N}(0, \sigma_r^2)$ is the additive white Gaussian noise, $\mathbf{p}^W_{a_l}$ is the anchor position, and $b_{a_l}$ is the bias associated with measurements from anchor $a_l$, respectively. In contrast to previous works, we model the biases as part of the same factor graph as the robot trajectory. The motivation for such a model is to capture the correlations between the robot trajectory and the measurement biases. Although biases are assumed to be constant in a particular window of the smoother, they can change between subsequent windows to accommodate spatially varying biases. A binary factor $\phi_{r_{lt}}$ between the latent robot state and the latent anchor-specific bias is added for every range measurement. To filter outliers, we employ a simple Euclidean distance check where range measurements that exceed the predicted range by $\delta r_{thr}$ are discarded. 

As mentioned previously, the update rate of VIO is generally higher than that of UWB. Consequently, multiple odometry estimates are received between two subsequent range measurements. The conventional approach of adding a node in the factor graph for every odometry estimate can result in a large graph size and increase the computational cost quickly. Alternatively, only those odometry estimates that are \textit{time-aligned} with the range measurements can be included in the graph \cite{Mascaro2018}, however, in doing so, useful odometry information is discarded. In order to include all of the odometry information while keeping the computational cost low, we take inspiration from IMU preintegration \cite{imupreint} and propose a similar method to integrate multiple odometry estimates into a single preintegrated measurement. 

\textit{Odometry preintegration}: We represent uncertainty in 3D poses using the formulation of \cite{Barfoot2014} but with the \textit{right perturbation} convention. Specifically, a generic pose is decomposed into a nominal pose $\bar{\mathbf{T}}_{i_t} \in  {SE}(3)$ and a perturbation $\perturb_t \in \mathbb{R}^{6 \times 1}$:
\begin{equation}
\mathbf{T}_{i_t} = \nompose_{i_t} \exp(\perturb_t^\wedge),
\label{eqn:pose_rep}
\end{equation}
where $\exp$ is a \textit{retraction} operation for $SE(3)$, $\wedge$ maps an element of $\mathbb{R}^6$ to an element of the associated \textit{Lie algebra} $\mathfrak{s}e(3)$, and $\perturb_t \sim \mathcal{N}(\mathbf{0}, \mathbf{\Sigma}_{i_t})$ is a Gaussian random variable. Consider a sequence of monotonic time steps $(t_0, t_1,..., t_k)$, and the corresponding odometry estimates obtained at those time steps: $\{\{\nompose^o_{i_1}, \mathbf{\Sigma}_{i_1}\}, \{\nompose^o_{i_2}, \mathbf{\Sigma}_{i_2}\},..., \{ \nompose^o_{i_k}, \mathbf{\Sigma}_{i_k}\}\}$. Here, $\nompose^o_{i_t} \in {SE}(3)$ represents the pose of the robot frame $\{i\}$ in the odometry frame $\{o\}$ at time $t$ and $\mathbf{\Sigma}_{i_t}$ is a positive definite matrix representing the corresponding covariance. First, we convert the odometry estimates into \textit{incremental} odometry between consecutive time steps: $\{ \{ \Delta \nompose^{i_0}_{i_{1}}, \mathbf{\Sigma}^\prime_{i_{1}} \}, \{ \Delta \nompose^{i_1}_{i_{2}}, \mathbf{\Sigma}^\prime_{i_{2}} \},, ...,\{ \Delta \nompose^{i_{k-1}}_{i_{k}}, \mathbf{\Sigma}^\prime_{i_{k}} \} \}$, where
\begin{align}
\Delta \nompose^{i_t}_{i_{t+1}} &=  (\nompose^o_{i_t})^{-1} \nompose^o_{i_{t+1}}, \\
\mathbf{\Sigma}^\prime_{i_{t+1}} &= \textbf{Ad}_{(\Delta \nompose^{i_t}_{i_{t+1}})^{-1}} \mathbf{\Sigma}_{i_{t}} \textbf{Ad}_{(\Delta \nompose^{i_t}_{i_{t+1}})^{-1}}^T + \mathbf{\Sigma}_{i_{t+1}},
\end{align}
and, $\textbf{Ad}_\pose$ is the \textit{Adjoint} operator for $SE(3)$. The incremental odometry estimates are then composed to give a single preintegrated measurement:
\begin{align}
\Delta \nompose^{i_t}_{i_{t+2}} &= \Delta \nompose^{i_t}_{i_{t+1}} \Delta \nompose^{i_{t+1}}_{i_{t+2}}, \label{eqn:comp_mean}\\
\mathbf{\Sigma}^{\prime\prime}_{i_{t+2}} &= \textbf{Ad}_{( \Delta \nompose^{i_{t+1}}_{i_{t+2}})^{-1}} \mathbf{\Sigma}^\prime_{i_{t+1}} \textbf{Ad}_{(\Delta \nompose^{i_{t+1}}_{i_{t+2}})^{-1}}^T + \mathbf{\Sigma^\prime}_{i_{t+2}}. \label{eqn:comp_cov}
\end{align}
The derivation of the above equations can be found in Appendix \ref{appn:odom_preint}. Equations \eqref{eqn:comp_mean} and \eqref{eqn:comp_cov} provide the mean and the covariance of the preintegrated odometry measurement which captures the total \textit{relative} displacement and orientation change as given the individual odometry estimates. Each preintegrated odometry measurement $\phi_{\mathbf{o}_t}$ is added as a binary factor between consecutive nodes in the factor graph as shown in Figure \ref{fig:smoother_factor_graph}. We estimate the latent states using \textit{maximum a posteriori} (MAP) inference. 

\subsection{WNOA-based fixed-lag smoother}

\begin{figure}[t]
	\center
	\begin{tikzpicture}[
	node distance={9mm},
	vertex/.style={circle, draw=black!100, thick, minimum width=0.85cm},
	empty_factor/.style={minimum size=1mm},
	]
	\node[circle,fill=black,inner sep=0pt,minimum size=5pt] (prior) at (0,0) {};
	\node[vertex] (x0) [right of=prior] {$\mathbf{p}^W_{o_0}$};
	\node[circle, draw, right of=x0, fill=black, inner sep=0pt,minimum size=5pt, label=below:{$\phi_{\mathbf{w}_0}$}] (o0){};
	\node[vertex] (x1)  [right of=o0] {$\mathbf{p}^W_{o_1}$};
	\node[circle, draw, right of=x1, fill=black, inner sep=0pt,minimum size=5pt, label=below:{$\phi_{\mathbf{w}_1}$}] (o1){};
	\node[vertex] (x2)  [right of=o1] {$\mathbf{p}^W_{o_2}$};
	\node[empty_factor] (o2) [right of=x2] {$\hdots$};
	\node[vertex] (xM) [right of=o2] {$\mathbf{p}^W_{o_k}$};
	\node[circle, draw, above of=x1, fill=black, inner sep=0pt,minimum size=5pt, label=left:{$\phi_{\mathbf{y}_{1}}$}] (y1){};
	\node[circle, draw, above of=x2, fill=black, inner sep=0pt,minimum size=5pt, label=right:{$\phi_{\mathbf{y}_{2}}$}] (y2){};
	\node[circle, draw, above of=xM, fill=black, inner sep=0pt,minimum size=5pt, label=right:{$\phi_{\mathbf{y}_{k}}$}] (yM){};
	%
	
	%
	\draw (prior) -- (x0);	
	\draw (x0) -- (o0);
	\draw (o0) -- (x1);
	\draw (x1) -- (o1);
	\draw (o1) -- (x2);
	\draw (x2) -- (o2);
	\draw (o2) -- (xM);
	\draw (x1) -- (y1);
	\draw (x2) -- (y2);
	\draw (xM) -- (yM);
	%
	\end{tikzpicture}
		\caption{Factor graph for the white-noise-on-acceleration (WNOA) fixed-lag smoother (WFLS). Motion priors obtained using the WNOA motion model are added as binary factors $\phi_{\mathbf{w}_t}$ to constrain the position component of the frame offset, $\mathbf{p}^W_{o_t}$. Frame offsets computed by UFLS are added as unary measurement factors $\phi_{\mathbf{y}_t}$.}
	\label{fig:wnoa_factor_graph}
	\vspace{-1em}
\end{figure}
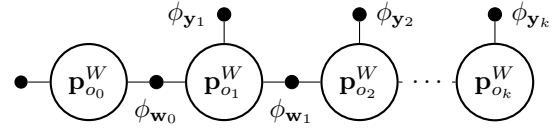

The output of UFLS is combined with a time-aligned odometry estimate to estimate the frame offset using \eqref{eqn:frame_offset}. The WNOA-based fixed-lag smoother (WFLS) uses a WNOA motion model and noisy estimates of the frame offset, ${\mathbf{T}}^W_o$, to generate smooth estimates of the frame offset, $\hat{\mathbf{T}}^W_o$. The factor graph for this setup is shown in Figure \ref{fig:sys_arch}. Generally, range measurements have a more direct impact on the position component ${\mathbf{p}}^W_o$ of the frame offset compared to the rotation component ${\mathbf{R}}^W_o$. This is because a range measurement influences the rotation component only through the lever arm, $\mathbf{p}^i_u$, which is generally small compared to the magnitude of range measurement. Hence, we impart a motion model only on the position component.

\textit{Motion model}: We adopt the Gaussian process (GP)-based continuous-time formulation proposed in \cite{Barfoot2014a} and impart the following linear time-invariant motion model on the position component $\mathbf{p}^W_{o_t}$:
\begin{align*}
	\dot{\mathbf{x}}(t) = \mathbf{A} \mathbf{x}(t) + \mathbf{L} \bm{\eta}_\mathbf{w}(t),
\end{align*}
with 
\begin{align*}
	\mathbf{x}(t) = \begin{bmatrix}
	{\mathbf{p}^W_{o_t}} \\
	\dot{\mathbf{p}}^W_{o_t} 
	\end{bmatrix},
	\mathbf{A} = \begin{bmatrix}
	\mathbf{0} & \mathbf{I} \\
	\mathbf{0} & \mathbf{0}
	\end{bmatrix},
	\mathbf{L} = \begin{bmatrix}
	\mathbf{0} \\
	\mathbf{I}
	\end{bmatrix},
\end{align*}
where $\mathbf{I}$ and $\mathbf{0}$ are the identity and zero matrices of appropriate dimensions, $\dot{\mathbf{p}}^W_{o_t}  = {d{\mathbf{p}}^W_{o_t}}/{dt}$, and $\bm{\eta}_\mathbf{w}(t) \sim \mathcal{GP}(\mathbf{0}, \mathbf{Q}_\mathbf{w} \delta(t - t'))$ is white noise drawn from a zero-mean GP with power spectral density matrix $\mathbf{Q}_\mathbf{w}$. 

Consider a sequence of monotonic times $(t_0, t_1,...,t_k)$. The mean state at time $t_j$ is
\begin{equation}
\mathbf{x}_{t_j} = \bm{\Phi}(t_j, t_{j-1}) \mathbf{x}_{j-1}, \label{eqn:wnoa_mean}
\end{equation}
 with the state transition matrix
\begin{align*}
	\bm{\Phi}(t_j, t_{j-1}) = \begin{bmatrix} 
		\mathbf{I} &  \Delta t_{j:j-1} \mathbf{I} \\
		\mathbf{0} & \mathbf{I}
	\end{bmatrix},
\end{align*}
where $\Delta t_{j:j-1} = t_{j} - t_{j-1}$. The corresponding covariance matrix is
\begin{align}
	\mathbf{Q}_{j:j-1} = \begin{bmatrix}
	\frac{1}{3} \Delta t^3_{j:j-1} \mathbf{Q}_\mathbf{w} & \frac{1}{2} \Delta t^2_{j:j-1} \mathbf{Q}_\mathbf{w} \\
	\frac{1}{2} \Delta t^2_{j:j-1} \mathbf{Q}_\mathbf{w} & \Delta t_{j:j-1} \mathbf{Q}_\mathbf{w}
	\end{bmatrix}.
	\label{eqn:wnoa_cov}
\end{align} 
The motion prior given by \eqref{eqn:wnoa_mean} and \eqref{eqn:comp_cov} is added as a binary factor $\phi_{\mathbf{w}_{j-1}}$ in the factor graph (see Figure \ref{fig:wnoa_factor_graph}). The measurement in this case is the position component of the frame offset at time $t_j$ computed using \eqref{eqn:frame_offset}
\begin{align}
	\mathbf{y}_j = \mathbf{p}^W_{o_j} + \bm{\eta}_{{oj}},
\end{align}
where $\bm{\eta}_{{oj}} \sim \mathcal{N}(\mathbf{0}, {\mathbf{\Sigma}}_{{\mathbf{p}}^W_{o_j}})$ is the additive white Gaussian noise. The covariance ${\mathbf{\Sigma}}_{{\mathbf{p}}^W_{o_j}}$ is obtained from UFLS as the marginal covariance. We add a unary factor $\phi_{\mathbf{y}_j}$ to the factor graph upon the receipt of a new frame offset. Similar to UFLS, latent states in WFLS are inferred using MAP inference. As before, the window size is parameterized by a time duration $\delta t_\textrm{wfls}$. The window size and the update rate of WFLS are independent of UFLS update rate and can be run at a lower or higher rate depending on the application. The output of WFLS is a smooth estimate of the position component of the frame offset $\hat{\mathbf{p}}^W_{o_j}$. The rotation component of the frame offset $\hat{\mathbf{R}}^W_{o_j}$ is obtained by plugging in the most recent estimate from UFLS and a time-aligned VIO measurement in \eqref{eqn:frame_offset}. Although we have imposed a motion model only on the position component, the extension to include the rotation component is straightforward. 
\section{Experimental Results} \label{sec:results}

In this section, we demonstrate the effectiveness of our method through simulation and real experiments. In simulation, we present qualitative results which highlight the importance of the WNOA motion model on the frame offset. We defer results quantifying the efficacy and accuracy of our method to real world experiments.

\begin{figure}[t]
	\centering
	\hspace*{-1em}
	\includegraphics[scale=0.65]{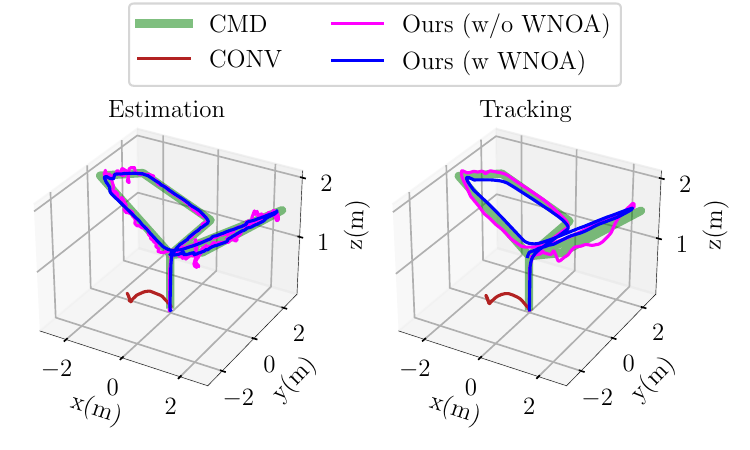}
	\caption{Trajectory estimation (left) and tracking (right) performance of different approaches. The inclusion of a white-noise-on-acceleration (WNOA) motion model  (Ours w WNOA) provides smooth corrections to VIO which results in reliable tracking of the commanded (CMD) trajectory compared to not including a WNOA motion model (Ours w/o WNOA) or the conventional approach (CONV) where output of UFLS is used for control.}
	\vspace*{-1.5em}
	\label{fig:sim_results}
\end{figure}

\begin{figure*}[t]
	\hspace*{-1.5em}
	\includegraphics[scale=0.6]{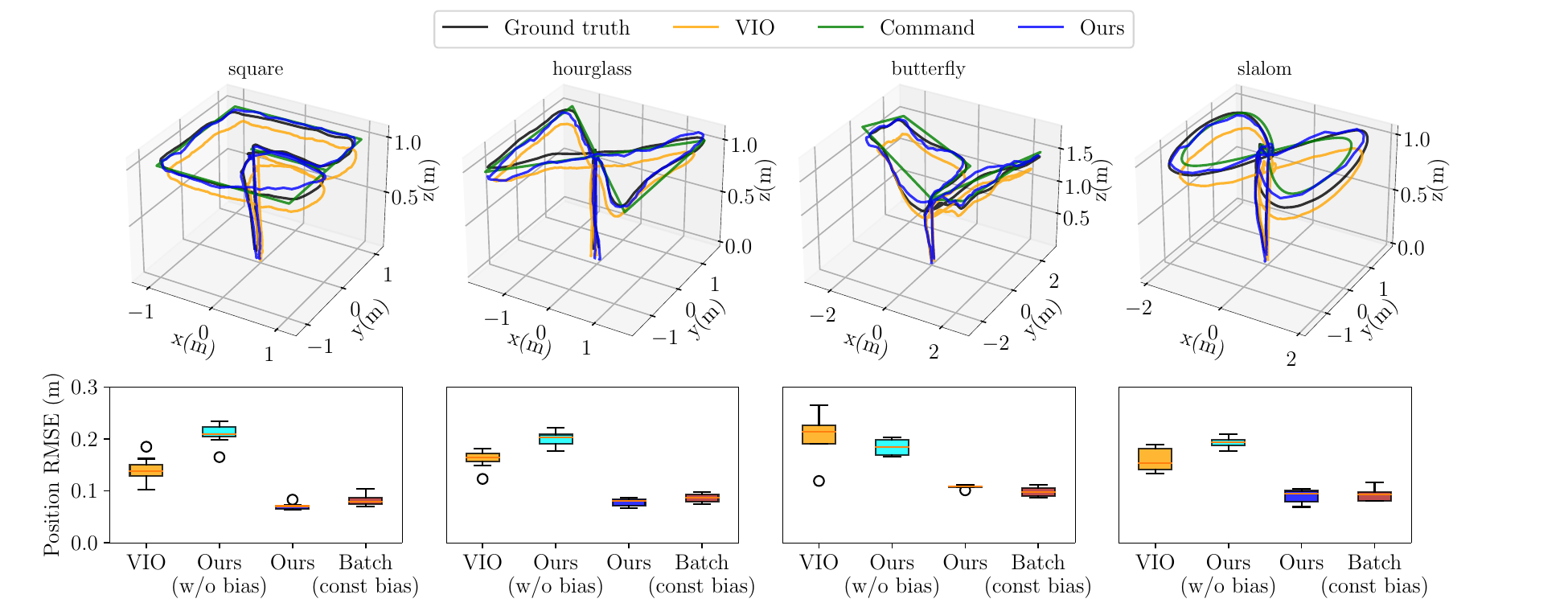}
	\caption{Results from closed-loop flight experiments at the UTIAS testbed. The top row shows the qualitative trajectory estimation performance. The ground truth trajectory also shows the closed-loop tracking performance compared to the commanded trajectory (Command). The bottom row shows the position root-mean-square error (RMSE) box plots for \textit{(i)}  visual-inertial odometry (VIO), \textit{(ii)} our method without anchor bias compensation (Ours w/o bias), \textit{(iii)} batch estimation with constant anchor bias compensation (Batch const bias), and \textit{(iv)} our method including anchor bias compensation (Ours). The maximum speed for the slalom path is $2\,\text{m/s}$ and for the others is $1\,\text{m/s}$.}
	\label{fig:vicon_results}
\end{figure*}

\begin{figure}[t]
	\centering
	\hspace*{-1em}\includegraphics[scale=0.7]{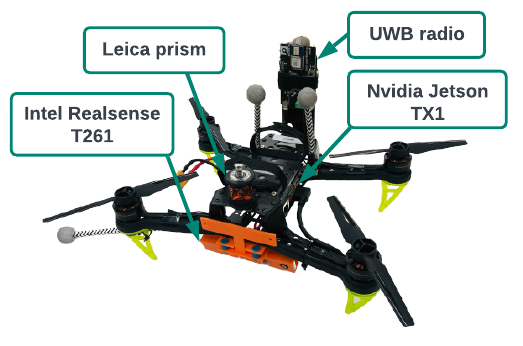}
	\caption{Our test platform used in experiments is a quadrotor equipped with a DW1000-based UWB radio and an Intel Realsense T261. All computation is performed onboard on a Nvidia Jetson TX1 computer. In areas where a motion capture system is not available, a Leica total station is used to get ground truth position by tracking the prism on the quadrotor.}
	\label{fig:barbary}
	\vspace*{-1.5em}
\end{figure}

\vspace*{-1em}
\subsection{Simulation}
We use the Gazebo simulator and a quadrotor from the Rotors simulator package as our test platform. The quadrotor is equipped with a generic range sensor and an odometry sensor. The noise parameters for the simulated sensors are chosen to reflect real sensor behavior. 

We perform a qualitative comparison of the closed-loop performance of our method against two different baselines in simulation as it provides a safe environment to test different methodologies. The first baseline is the conventional (CONV) method where the output of the fixed-lag smoother UFLS is used for closed-loop control. The second baseline uses the output of WFLS but without any smoothing, dubbed Ours (w/o WNOA). This is similar to the method of \cite{Mascaro2018}. We refer to our proposed method as Ours (w WNOA). The same minimum-jerk trajectory is commanded in all experiments. The stock proportional-integral-derivative (PID) controllers from the Rotors simulator package are used for both position and attitude control.

The results are shown in Figure \ref{fig:sim_results}. The CONV baseline results in unstable tracking due to the low rate and non-smooth output of UFLS. The second baseline tracks the commanded trajectory, however, the non-smooth nature of the estimated frame offset results in poor tracking performance. In contrast, our proposed method tracks the commanded trajectory reliably. We provide additional details on the benefit of the additional smoothing in Appendix \ref{appn:smooth_effect}.

%

\subsection{Real experiments}\label{sec:real-world-experiments}

To quantify the performance of the proposed method, we performed closed-loop flight experiments in three real indoor environments: \textit{(i)} University of Toronto, Institute for Aerospace Studies (UTIAS) testbed, \textit{(ii)} UTIAS cafeteria, and \textit{(iii)} Myhal testbed. Due to space constraints, we provide quantitative results from the UTIAS testbed and the UTIAS cafeteria. Footage of the flight experiments from all three environments can be found in the accompanying video\footnote{\url{http://tiny.cc/uwb_vio}}.

\textit{Baselines}: We focus on trajectory estimation accuracy, which is the core contribution of our work. We compare our method against three different baselines. The first baseline consists of VIO-only estimation, which we refer to as \textit{VIO}. Since VIO provides relative pose estimates, we initialize the VIO algorithm close to the ground truth only for comparison. The second baseline is similar to the method proposed in \cite{Mascaro2018} where no bias estimation is done, which we refer to as \textit{Ours (w/o bias)}. Note that we had to modify the method of \cite{Mascaro2018} to incorporate range measurements as the UWB system used in our experiments does not provide position measurements. The third baseline is batch trajectory estimation \cite[Chapter 4]{Barfoot2023}. This represents the ideal scenario where all of the sensor data is available for estimation. For batch estimation, we consider a single bias latent variable for each anchor across the entire trajectory, which we refer to as \textit{Batch (const bias)}. We refer to our method as \textit{Ours}.

\textit{Setup}: We use DW1000-based UWB radios from Bitcraze. The position of the UWB anchors is measured using a Leica total station. Our test platform is a custom-built quadrotor (see Figure \ref{fig:barbary}) equipped with a UWB radio which provides range measurements, and an Intel Realsense T261 that provides VIO. We disable the the loop closure feature on the T261 to run it in VIO-only mode. All computation is performed onboard on a Jetson TX1 single-board computer. The lower-level flight controller is run on a PX4-based autopilot with stock firmware. The quadrotor is also equipped with a Leica prism to obtain ground truth position in environments where a motion capture system is not available. The UWB radios are operated in two-way range (TWR) mode. The range measurement outlier rejection threshold is chosen as $\delta r_\textrm{thr} = 0.5\,\si{m}$, which is larger than $3\sigma_r$ with $\sigma_r = 0.1\,\si{m}$. The two fixed-lag smoothers, UFLS and WFLS, are implemented using the GTSAM library \cite{gtsam}. The frequency of range measurements and odometry from VIO are ~17\,Hz and ~200\,Hz, respectively. The update rate of UFLS is 5\,Hz and that of WFLS is 10\,Hz with $\delta t_\textrm{ufls} = 1\,\si{s}$ and $\delta t_\textrm{wfls} = 1\,\text{s}$.

\textit{Experimental procedure}: The proposed estimation and control pipelines run online on the onboard computer of the quadrotor. For navigation, we generate minimum-jerk trajectories using coarse user-defined waypoints. A trajectory sampler uses the estimated state to generate pose and velocity setpoints to track the reference minimum-jerk trajectory. In a new environment, we perform a single autonomous calibration flight to obtain coarse values for anchor-related biases, which are then used as priors for subsequent flights. In each experiment, estimation results for the proposed method along with the sensor data are recorded onboard. The recorded sensor data is used to evaluate the different baselines \textit{offline}.

\begin{figure}[t]
	\centering
	\includegraphics[scale=0.34]{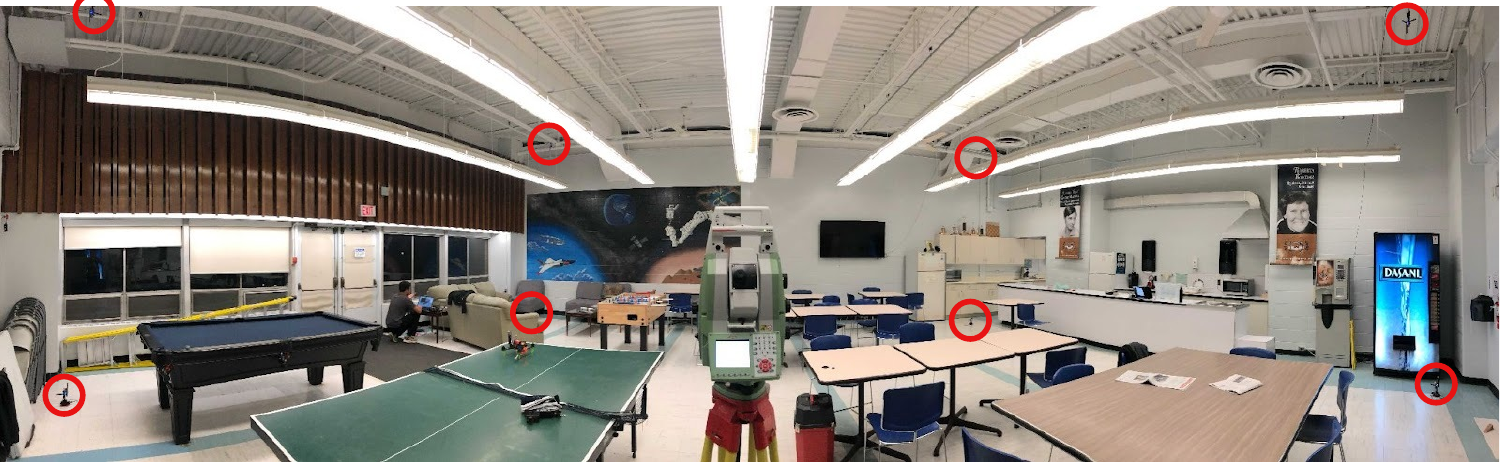}
	\caption{A panoramic view of the UTIAS cafeteria with the positions of the UWB anchors highlighted in red circles. The dimensions of the cafeteria are $10\,\si{m} \times 10\,\si{m} \times 5\,\si{m}$.}
	\label{fig:cafe}
\end{figure}

\begin{table}[b]
	\center
	\setlength\extrarowheight{2.5pt}
	\caption{Best-case trajectory tracking performance using the proposed method from flight tests in the UTIAS testbed.}
	\begin{tabular}{| c | c | c | c | c |} 
		\hline
		Path & square & hourglass & butterfly & slalom \\
		\hline
		RMS Tracking Error ($\si{m}$) & 0.106 & 0.118 & 0.135 & 0.232 \\
		\hline
	\end{tabular}
	\label{tab:tracking_performance}
\end{table}

\textit{UTIAS testbed}: Our first test space is an indoor testbed with 8 UWB anchors at the corners of a flight arena of dimensions $7\,\si{m} \times 8\,\si{m} \times 3.5\,\si{m}$ The arena is equipped with a Vicon motion capture system for ground truth. We performed multiple closed-loop flight experiments with 4 different paths shown in Figure \ref{fig:vicon_results}. The commanded speed for the first 3 paths is $1\,\si{m/s}$, whereas the commanded speed for the slalom path is $2\,\si{m/s}$. Box plots showing the absolute position root-mean-square error (RMSE) are provided in the bottom row of Figure \ref{fig:vicon_results}. The top row shows qualitative estimation and tracking results from one such experiment for the different paths. Note that since we are doing closed-loop flights, the ground truth (GT) trajectory shows the tracking performance. 

The box plots highlight the importance of anchor bias estimation as it doubles the estimation accuracy in many cases. Since batch estimation considers the full measurement history, it is expected to be better than the proposed method. However, in some cases, the proposed method is more accurate than batch estimation, which can be attributed to \textit{spatially} varying biases. Specifically, batch estimation considers constant anchor biases for the entire trajectory whereas the proposed method considers the biases to be constant only within the current estimation window. This observation is consistent with the results of \cite{Prorok2012}. In most cases we achieve a position RMSE below $10\,\si{cm}$. The performance on the butterfly trajectory is poorer as the MAV travels to the extremities of the test arena where the estimation is affected by the dilution of precision \cite[Section 7.3]{kaplan}. The position RMSE values observed in the square path are similar to the values observed in \cite{Fang2021}. The results are promising considering that the DW1000-based UWB radios used in our experiments have a precision of $\pm10\,\si{cm}$ compared to the P440 UWB radios used in \cite{Fang2021} which have a precision of $\pm1\,\si{cm}$. The results show that the assumption of constant bias within the current estimation window is sufficient to achieve (sub)decimeter-level accuracy in many cases. 

Although the core contribution of our work is on localization, we provide the best-case trajectory tracking performance for the four test trajectories in Table \ref{tab:tracking_performance}. The results show that our method achieves smooth and reliable tracking at moderate speeds with minimally tuned off-the-shelf open-source controllers. The higher tracking error in the slalom path can be attributed to the minimally tuned controller as the MAV overshoots the commanded trajectory at higher speeds. Note however that the trajectory estimation accuracy is still high in this case (see boxplot in Figure \ref{fig:vicon_results}). 


\textit{UTIAS cafeteria}: Our second test space is the UTIAS cafeteria. A panoramic view of the cafeteria is shown in Figure \ref{fig:cafe}. The dimensions of the cafeteria are $10\,\si{m} \times 10\,\si{m} \times 5\,\si{m}$ with 8 UWB anchors installed on the ceiling and on the floor. This is a very challenging setup for UWB-based localization as all of the anchors are occluded by metallic and non-metallic objects that induce serve NLOS scenarios.

\definecolor{firebrick}{rgb}{0.7, 0.13, 0.13}
\begin{table}[b]
	\center
	\setlength\extrarowheight{2.5pt}
	\caption{Position root-mean-square error (RMSE) for VIO-only estimation, batch trajectory estimation with constant bias compensation (Batch const bias) and our method (Ours) from experiments in UTIAS cafeteria.}
	\begin{tabular}{ | c | c | c | c | c |} 
		\hline
		\multirow{2}{*}{Algorithm} 
		& \multicolumn{4}{c|}{Position RMSE ($\si{m}$)}\\
		\cline{2-5}
		& Path 1 & Path 2 & Path 3 & Path 4 \\
		\hline
		\textcolor{orange}{VIO} & 0.272 & 0.434 & 0.208 & 0.227\\ 
		\hline
		\textcolor{firebrick}{Batch (const bias)} & \textbf{0.125} & 0.217 & \textbf{0.150} & 0.137 \\
		\hline
		\textcolor{blue}{Ours} & 0.130 & \textbf{0.154} & 0.152 & \textbf{0.121} \\
		\hline
	\end{tabular}
	\label{tab:cafe_estimation}
\end{table}

\begin{figure}[t]
	\hspace*{-0.5em}
	\includegraphics[scale=0.65]{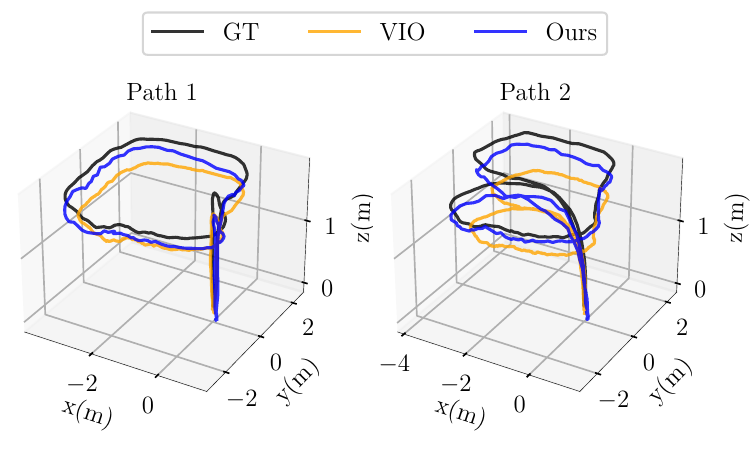}
	\caption{Trajectory estimation results from two experiments in the UTIAS cafeteria. The quadrotor is manually moved along different paths in the experiments. The proposed estimation pipeline is run online onboard the quadrotor and results from our method (Ours), VIO-only estimation (VIO), and ground truth (GT) are recorded. The corresponding position root-mean-square error (RMSE) values are provided in Table \ref{tab:cafe_estimation}.}
	\label{fig:cafe_estimation}
\end{figure}

\begin{figure}[t]
	\hspace*{-0.5em}
	\includegraphics[scale=0.65]{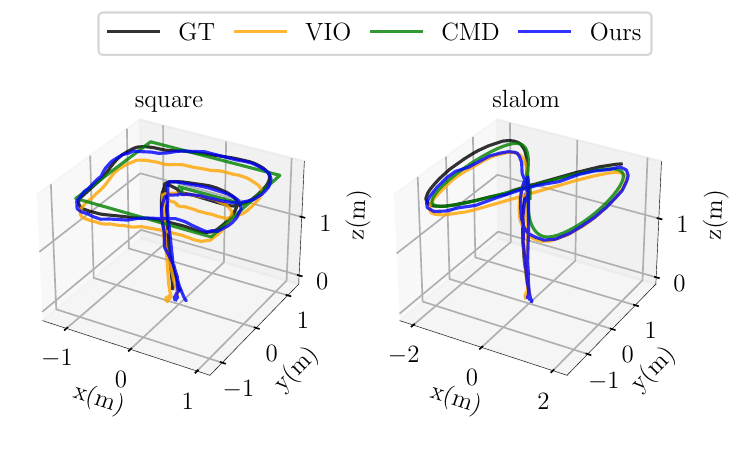}
	\caption{Results from closed-loop flight experiments in the UTIAS cafeteria. The estimated trajectories from VIO-only estimation (VIO) and our proposed method (Ours) are shown. The maximum speeds for the square path and the slalom path are $1\,\si{m/s}$ and $2\,\si{m/s}$, respectively. 
	}
	\label{fig:cafe_tracking}
\end{figure}
For ground truth data, we use a Leica total station in tracking mode. In this mode, the total station tracks a prism that is mounted on the quadrotor (see Figure \ref{fig:barbary}) and outputs position of the prism at $5\,\si{Hz}$. We observed that the total station would lose track of the quadrotor during closed-loop flights. Hence, to quantify the estimation accuracy we simulated closed-loop flights by moving the quadrotor manually at lower speeds. In each of the experiments, the estimation pipeline runs online onboard the quadrotor. Results from two such experiments are shown in Figure \ref{fig:cafe_estimation}. The corresponding position RMSE values from four experiments are provided in Table \ref{tab:cafe_estimation}. The accuracy of the proposed method is lower compared to the previous case but similar to that observed with the butterfly path in the UTIAS testbed. We see that in some cases the proposed method performs better than the batch approach, which can be attributed to spatially varying biases.

We performed flight experiments with the same four paths from the previous setup. Results from two such experiments are shown in Figure \ref{fig:cafe_tracking}. During the experiments, we observed that most of the anchors were occluded and less than half were in line of sight at any given time. The proposed method is still able to accurately track the desired trajectories at moderate speeds in highly-cluttered environments.

\section{Conclusion and Future work} \label{sec:conclusion}
In this paper, we presented a dual fixed-lag smoother approach to fusion of range measurements and VIO for localization and navigation of MAVs. We showed that the proposed method achieves accurate localization and also generates smooth pose estimates for control. The proposed approach is able to estimate any systematic biases in range measurements resulting in a two-fold increase in localization accuracy. Our method is lightweight and can run on a computationally constrained MAV. Through multiple real world experiments, we showed that the proposed approach achieves decimeter-to-sub-decimeter-level localization accuracy with off-the-shelf sensors and open-source controllers in cluttered indoor environments.


\bibliographystyle{unsrt}
\bibliography{main.bib}

\newpage

\appendix
This appendix accompanies the arXiv verison of
this paper.

\subsection{Odometry preintegration}\label{appn:odom_preint}

\begin{figure}[b]
	\centering
	\includegraphics[scale=0.55]{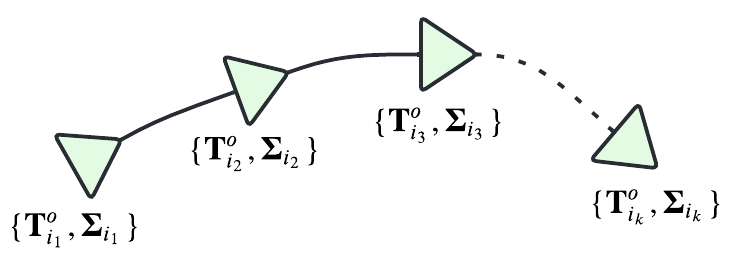}
	\caption{Visual-inertial odometry (VIO) estimates the robot pose $\mathbf{T}^o_{i_t}$ at different times steps along with the corresponding uncertainty $\mathbf{\Sigma}_{i_t}$. In odometry preintegration multiple odometry measurements are summarized with a single preintegrated odometry measurement.}
	\label{fig:odometry}
\end{figure}

\begin{figure}[t]
	\begin{subfigure}[t]{\textwidth}
		\includegraphics[scale=0.75,trim={0cm 1cm 1cm 0cm},clip]{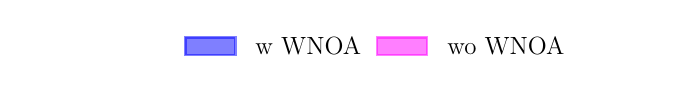}
	\end{subfigure}
	\begin{subfigure}[t]{0.5\textwidth}
		\centering
		\includegraphics[scale=0.9,trim={0.3cm 0 0cm 0cm},clip]{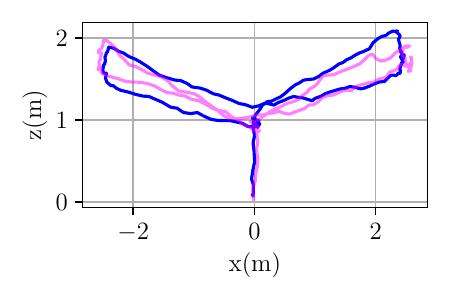}
		\caption{Output of first fixed-lag smoother (UFLS).}
		\label{fig:wnoa_res_est}
	\end{subfigure}
	\begin{subfigure}[t]{0.5\textwidth}
		\centering
		\includegraphics[scale=0.9,trim={0.42cm 0 0 0},clip]{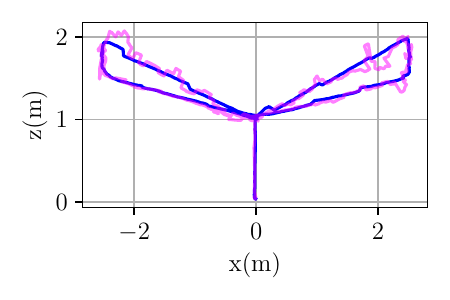}
		\caption{Output of second fixed-lag smoother (WFLS) used for closed-loop control.}
		\label{fig:wnoa_res_smo}
	\end{subfigure}
	\caption{Trajectory tracking performance with (w WNOA) and without (wo WNOA) additional smoothing. (a) The trajectory estimated by the first fixed-lag smoother (UFLS). (b) Output of the second fixed-lag smoother used for close-loop control.}
	\label{fig:wnoa_res}
\end{figure}

We represent the uncertainty associated with 3D poses on the corresponding Lie algebra \cite{Barfoot2023}. Specifically, a pose $\mathbf{T}_{i_t} \in {SE}(3)$ is decomposed into a nominal pose $\bar{\mathbf{T}}_{i_t} \in  {SE}(3)$ and a small perturbation $\perturb_t \in \mathbb{R}^{6 \times 1}$:
\begin{equation}
\mathbf{T}_{i_t} = \nompose_{i_t} \exp(\perturb_t^\wedge),
\label{eqn:pose_rep}
\end{equation}
where $\exp$ is the retraction operation for $SE(3)$, and $\wedge$ is the \textit{hat} operator that maps an element of $\mathbb{R}^6$ to an element of the Lie algebra $\mathfrak{s}e(3)$, and $\perturb_t \sim \mathcal{N}(\mathbf{0}, \mathbf{\Sigma}_{i_t})$ is a Gaussian random variable. Consider a sequence of monotonic time steps $(t_0, t_1,..., t_k)$ and the corresponding odometry estimates $\{\{\nompose^o_{i_1}, \mathbf{\Sigma}_{i_1}\}, \{\nompose^o_{i_2}, \mathbf{\Sigma}_{i_2}\},..., \{ \nompose^o_{i_k}, \mathbf{\Sigma}_{i_k}\}\}$ as shown in Figure \ref{fig:odometry}. Here $\nompose^o_{i_t} \in {SE}(3)$ represents the pose of the robot frame $\{i\}$ with respect to the odometry frame $\{o\}$ at time $t$ and $\mathbf{\Sigma}_{i_t}$ is a positive definite matrix representing the corresponding covariance. First, we convert the absolute odometry measurements into incremental odometry between consecutive time steps:
\begin{align}
\Delta \pose^{i_t}_{i_{t+1}} &= (\pose^o_{i_t})^{-1} \pose^o_{i_{t+1}}, \nonumber \\
&= (\nompose^o_{i_t} \exp(\perturb_t^\wedge) )^{-1} \nompose^o_{i_{t+1}} \exp(\perturb_{t+1}^\wedge), \nonumber \\
&= \exp(-\perturb_t^\wedge) \underbrace{(\nompose^o_{i_t})^{-1} \nompose^o_{i_{t+1}}}_{\Delta\nompose^{i_t}_{i_{t+1}}} \exp(\perturb_{t+1}^\wedge), \nonumber \\
%
%
%
%
&= \Delta \nompose^{i_t}_{i_{t+1}} \exp ( - (\textbf{Ad}_{(\Delta \nompose^{i_t}_{i_{t+1}})^{-1}} \perturb_t ) ^\wedge )  \exp(\perturb_{t+1}^\wedge), \nonumber
\end{align}
where $\textbf{Ad}_{()}$ is the \textit{Adjoint} operator for $SE(3)$. Thus, the incremental odometry can be represented as combination of a nominal pose: 
\begin{equation}
\Delta \nompose^{i_t}_{i_{t+1}} =  (\nompose^o_{i_t})^{-1} \nompose^o_{i_{t+1}}
\label{eqn:inc_mean}
\end{equation}
and a perturbation:
\begin{align*}
\exp(\perturb_{t+1}^{\prime^\wedge}) = \exp ( -(\textbf{Ad}_{(\Delta \nompose^{i_t}_{i_{t+1}})^{-1}} \perturb_t ) ^\wedge )  \exp(\perturb_{t+1}^\wedge).
\end{align*}
To calculate the covariance associated with incremental odometry, we apply the \textit{Baker–Campbell–Hausdorff} (BCH) formula to the above equation and use a second-order approximation \cite{Barfoot2023}:
\begin{equation}
\mathbf{\Sigma}^\prime_{i_{t+1}} = \textbf{Ad}_{(\Delta \nompose^{i_t}_{i_{t+1}})^{-1}} \mathbf{\Sigma}_{i_{t}} \textbf{Ad}_{(\Delta \nompose^{i_t}_{i_{t+1}})^{-1}}^T + \mathbf{\Sigma}_{i_{t+1}}
\label{eqn:inc_cov}
\end{equation}
Using \eqref{eqn:inc_mean} and \eqref{eqn:inc_cov}, we can convert the odometry measurements into a sequence of incremental odometry measurements $\{ \{ \Delta \nompose^{i_0}_{i_{1}}, \mathbf{\Sigma}^\prime_{i_{1}} \}, \{ \Delta \nompose^{i_1}_{i_{2}}, \mathbf{\Sigma}^\prime_{i_{2}} \},, ...,\{ \Delta \nompose^{i_{k-1}}_{i_{k}}, \mathbf{\Sigma}^\prime_{i_{k}} \} \}$. 

The incremental odometry measurements can be composed recursively as follows.
\begin{align*}
\Delta \pose^{i_t}_{i_{t+2}} &= \Delta \pose^{i_t}_{i_{t+1}} \Delta \pose^{i_{t+1}}_{i_{t+2}},\\
&= \Delta \nompose^{i_t}_{i_{t+1}} \exp(\perturb_{t+1}^{\prime^\wedge}) \Delta \nompose^{i_{t+1}}_{i_{t+2}} \exp(\perturb_{t+2}^{\prime^\wedge}), \\
%
%
%
%
&= \Delta \nompose^{i_t}_{i_{t+1}} \Delta \nompose^{i_{t+1}}_{i_{t+2}} \exp ( \perturb_{t+1}^{\prime\prime ^\wedge} ) \exp(\perturb_{t+2}^{\prime^\wedge}),
\end{align*}
where $\perturb_{t+1}^{\prime\prime} = \textbf{Ad}_{(\nompose^{i_{t+1}}_{i_{t+2}})^{-1}} \perturb_{t+1}^{\prime}$. The mean and covariance of the composition can be derived similar to \eqref{eqn:inc_mean} and \eqref{eqn:inc_cov}:
\begin{align}
\Delta \nompose^{i_t}_{i_{t+2}} &= \Delta \nompose^{i_t}_{i_{t+1}} \Delta \nompose^{i_{t+1}}_{i_{t+2}}, \label{eqn:comp_mean}\\
\mathbf{\Sigma}^{\prime\prime}_{i_{t+2}} &= \textbf{Ad}_{( \Delta \nompose^{i_{t+1}}_{i_{t+2}})^{-1}} \mathbf{\Sigma}^\prime_{i_{t+1}} \textbf{Ad}_{(\Delta \nompose^{i_{t+1}}_{i_{t+2}})^{-1}}^T + \mathbf{\Sigma^\prime}_{i_{t+2}}. \label{eqn:comp_cov}
\end{align}

Equations \eqref{eqn:comp_mean} and \eqref{eqn:comp_cov} provide the mean and the covariance associated with the preintegrated odometry measurement.

\subsection{Effect of WNOA smoothing} \label{appn:smooth_effect}
In this section, we provide additional details to highlight the benefit of additional smoothing for online control.

As mentioned previously, we decouple the core state estimation routine and the localization routine that is used for control. Specifically, the core state estimation is carried out using a fixed-lag smoother (UFLS) that combines range measurement and odometry and does not impose any smoothness requirements. The output of UFLS will be non-smooth on account of corrections from range measurements and hence is not used directly for close-loop control (see Figure \ref{fig:wnoa_res_est}).

The output of UFLS is then fed to a second smoother (WFLS) which imposes smoothness requirements using a white-noise-on-acceleration (WNOA) motion model. Note that the output of this particular smoother is used purely for control and does not influence the core state estimation routine or the localization RMSE. This additional smoothing could result in lower trajectory tracking accuracy but we observed empirically that without additional smoothing the trajectory tracking accuracy is lower and results in unstable flights as shown in Figure \ref{fig:wnoa_res_smo}.


\end{document}